\documentclass[conference]{IEEEtran}
\IEEEoverridecommandlockouts
\makeatletter
\def\ps@IEEEtitlepagestyle{%
  \def\@oddfoot{\mycopyrightnotice}%
  \def\@evenfoot{}%
}
\def\mycopyrightnotice{%
  {\footnotesize {\itshape Accepted to IEEE Conference on Games 2021:} 978-1-6654-3886-5/21/\$31.00 \copyright 2021 IEEE\hfill}% <--- Change here
  \gdef\mycopyrightnotice{}% just in case
}
\usepackage{flushend}
\usepackage{cite}
\usepackage{amsmath,amssymb,amsfonts}
\usepackage{algorithmic}
\usepackage{graphicx}
\usepackage{textcomp}
\usepackage{xcolor}
\usepackage{url}
\def\BibTeX{{\rm B\kern-.05em{\sc i\kern-.025em b}\kern-.08em
    T\kern-.1667em\lower.7ex\hbox{E}\kern-.125emX}}
\begin{document}

\title{Carle's Game: An Open-Ended Challenge in Exploratory Machine Creativity\\
}

\author{\IEEEauthorblockN{1\textsuperscript{st} Q. Tyrell Davis}
\IEEEauthorblockA{\textit{} \\
%\textit{}\\
%Austin, Texas, USA\\
Wyoming, USA
qtyrelldavis@protonmail.com}
%\and
}

\maketitle

\begin{abstract}

	This paper is both an introduction and an invitation. It is an introduction to CARLE, a Life-like cellular automata simulator and reinforcement learning environment. It is also an invitation to Carle's Game, a challenge in open-ended machine exploration and creativity. Inducing machine agents to excel at creating interesting patterns across multiple cellular automata universes is a substantial challenge, and approaching this challenge is likely to require contributions from the fields of artificial life, AI, machine learning, and complexity, at multiple levels of interest. Carle's Game is based on machine agent interaction with CARLE, a Cellular Automata Reinforcement Learning Environment. CARLE is flexible, capable of simulating any of the 262,144 different rules defining Life-like cellular automaton universes. CARLE is also fast and can simulate automata universes at a rate of tens of thousands of steps per second through a combination of vectorization and GPU acceleration. Finally, CARLE is simple. Compared to high-fidelity physics simulators and video games designed for human players, CARLE's two-dimensional grid world offers a discrete, deterministic, and atomic universal playground, despite its complexity. In combination with CARLE, Carle's Game offers an initial set of agent policies, learning and meta-learning algorithms, and reward wrappers that can be tailored to encourage exploration or specific tasks. 
	
\end{abstract}


\section{Introduction}
	
% fluffy, philosophical shtick about exploring the universe and the implied loneliness of the anthropic principle

	Intelligence is an emergent phenomenon that requires nothing more than the presence of matter and energy, the physical constraints of our present universe, and time. At the very least we know that the preceding statement is true by an existence proof, one that continues to demonstrate itself in the very act of the reader parsing it. There may be simpler ways to generate intelligence, and there are certainly paths to intelligence that are significantly more complex, but the only path that we have positive proof for so far is the evolution of life by natural selection. Put simply, an algorithm that can be described as ``the most likely things to persist will become more likely to persist" is at least capable of creating intelligence --- though the likelihood of this happening is unknown and shrouded in uncertainty \cite{kipping2020} \footnote{While I've often read that human existence is the result of a ``a single run" of an open-ended process \cite{stanley2019}, we have no way to no for certain how many similar ``runs" may have preceded (or will succeed) the experience of our own perspective. This is the anthropic principle in a nutshell \cite{carter1974}}. Consequently the study of intelligence, both natural and artificial, from a viewpoint informed by open-ended complexity is not only an alternative approach to building artificially intelligent systems, but a way to better understand the context of Earth-based intelligence as it fits into the larger universe.
	
	One hallmark of modern approaches to artificial intelligence (AI), particularly in a reinforcement learning framework, is the ability of learning agents to exploit unintended loopholes in the way a task is specified \cite{amodei2016}. Known as specification gaming or reward hacking, this tendency constitutes the crux of the control problem and places an additional engineering burden of designing an ``un-hackable" reward function, the substantial effort of which may defeat the benefits of end-to-end learning \cite{singh2019}. An attractive alternative to manually specifying robust and un-exploitable rewards is to instead develop methods for intrinsic or learned rewards \cite{burda2018}. For tasks requiring complex exploration such as the ``RL-hard" Atari games Montezuma's Revenge and Pitfall, intrinsic exploration-driven reward systems are often necessary \cite{ecoffet2019}. 

	Games have a long history as a testing and demonstration ground for artificial intelligence. From MENACE, a statistical learning tic-tac-toe engine from the early 1960s \cite{gardner1962}, to the world champion level chess play of Deep Blue \cite{tan1995} and the lesser known Chinook's prowess at checkers \cite{schaeffer2007} in the 1990s, to the more recent demonstrations of the AlphaGo lineage \cite{silver2016, silver2017, silver2018, schrittwieser2020} and video game players like AlphaStar for StarCraft II \cite{vinyals2019} and OpenAI Five for Dota 2 \cite{berner2019} to name just a few. Games in general provide an opportunity to develop and demonstrate expert intellectual prowess, and consequently have long been an attractive challenge for AI. While impressive, even mastering several difficult games is but a small sliver of commonplace human and animal intelligence. The following statement may be debatable to some, but mastering the game of Go is not even the most interesting demonstration of intelligence one can accomplish with a Go board. 

	While taking breaks from more serious mathematical work, John H. Conway developed his Game of Life by playing with stones on a game board \cite{gardner1970}. The resulting Game demonstrated a system of simple rules that would later be proven to be computationally universal. Conway's Life did not found the field of cellular automata (CA), but it did motivate subsequent generations of research and play and is likely the most well known example of CA.

\section{Life-like cellular automata crash-course}

The Game of Life was invented by John H. Conway through mathematical play in the late 1960s, and it was introduced to the public in a column in Martin Gardner's ``Mathematical Games" in Scientific American in 1970 \cite{gardner1970}. The Game was immensely influential and inspired subsequent works in similar systems and cellular automata in general. Conway's Life demonstrated the emergence of complex behavior from simple rules and would later be proven a Turing complete system capable of universal computation \cite{rendell2014}. 

%Conway was somewhat by the popular appreciation of Life throughout most of his life, preferring that 
\begin{figure}[tb]
	\centerline{\includegraphics[scale=0.3]{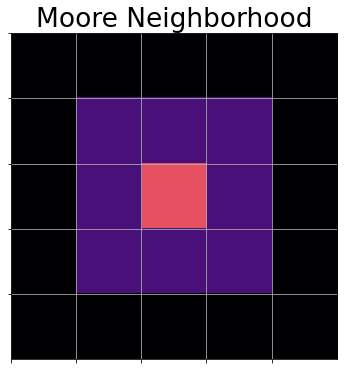}}
\caption{A Moore neighborhood in purple surrounding a cell in pink. }
\label{fig:moore}
\end{figure}

Conway's Life is based on the changing binary states of a 2-dimensional grid. Cell units in the grid change their state based on local information, determined solely by the sum of states of their immediate neighbors. The 3 by 3 grid surrounding each cell (excluding itself) is called a Moore neighborhood (Figure \ref{fig:moore}). Each cell's next state is fully determined by its current state and the sum of cell states in its Moore neighborhood. Cellular Automata (CA) based on this update configuration are known as ``Life-like". Different Life-like CA can be described by the rules they follow for transitioning from a state of 0 to a state of 1 (``birth") and maintaining a state of 1 (``survival"). Life-like rules can be written in a standard string format, with birth conditions specified by numbers preceded by the letter ``B" and survival conditions by numbers preceded by ``S", and birth and survival conditions separated by a slash. For example, the rules for Conway's Life are B3/S23, specifying the dead cells with 3 live neighbors become alive (state 1) and live cells with 2 or 3 live neighbors remain alive, all other cells will remain or transition to a dead state (state 0). 

\begin{figure}[tb]
	\centerline{\includegraphics[scale=0.25]{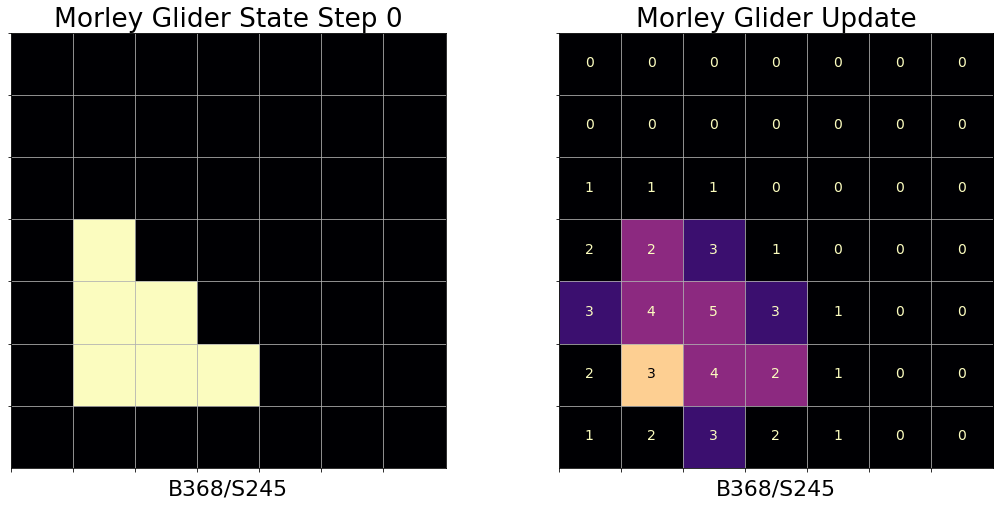}}
	\centerline{\includegraphics[scale=0.25]{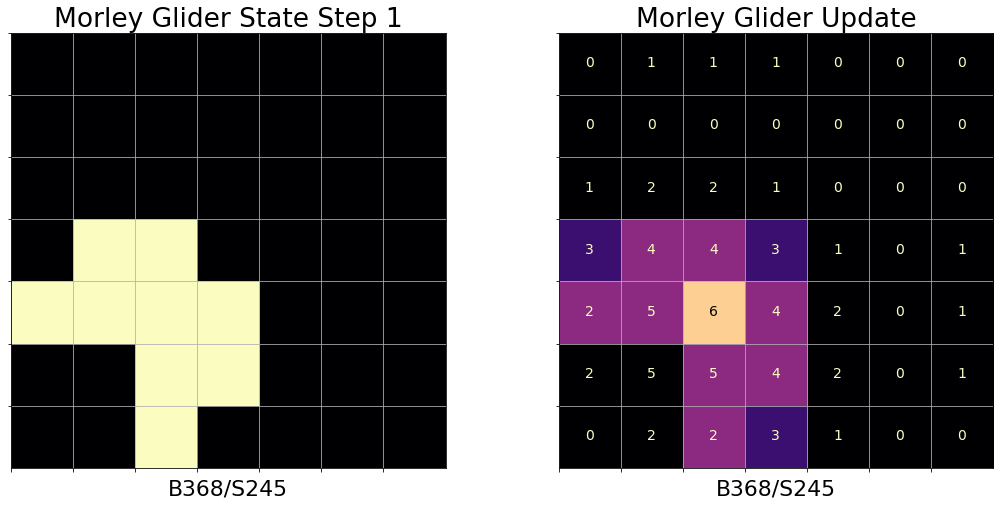}}
 	\centerline{\includegraphics[scale=0.25]{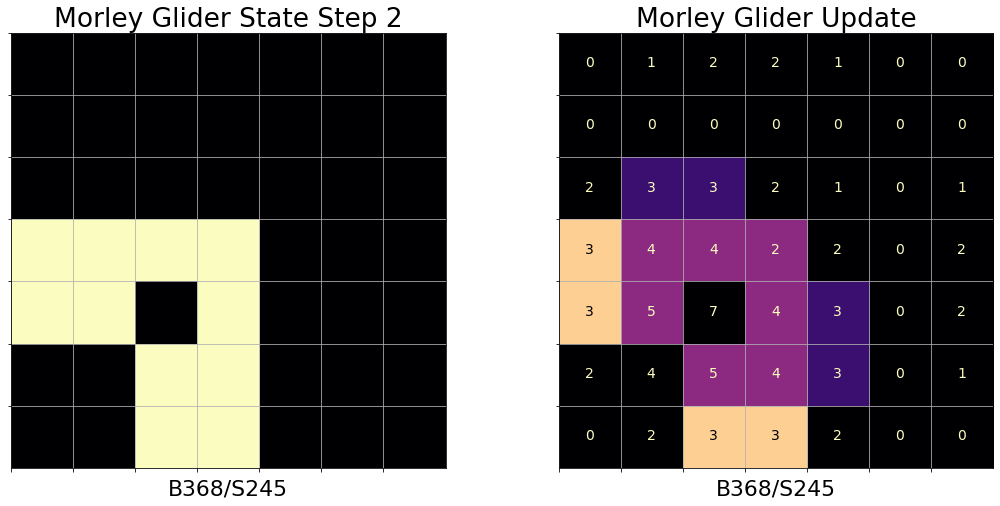}}
% 	\centerline{\includegraphics[scale=0.25]{morley_glider_3.png}}
\caption{The first 2 updates of a glider pattern in the B368/S245 set of rules. The Moore neighborhood values are inset in each cell, with birth and survival denoted in purple and ``death" transitions denoted in pinkish yellow. Cells with a starting state of 0 and 3, 6, or 8 neighbors transition to a state of 1, while cells with a state of 1 with 2, 4, or 5 neighbors remain in a state of 1. All other cells transition to a state of 0.}
\label{fig:morley0}
\end{figure}

Including Conway's Life (B3/S23), there are $2^9 * 2^9 = 262,144$ possible rule variations in Life-like CA. Many of these rules support objects reminiscent of machine or biological entities in their behavior. Types of CA objects of particular interest include spaceships, objects that translocate across the CA grid as they repeat several states; puffers, spaceships that leave a path of persistent debris behind them, computational components like logic gates and memory, and many more. The update progression of a glider pattern under B368/S245 rules, also known as Morley or Move, is shown in Figure \ref{fig:morley0}. An active community continues to build new patterns and demonstrations of universal computation in Life-like CA \cite{conwaylifeforum}.

%\begin{figure}[tb]
% 	\centerline{\includegraphics[scale=0.25]{morley_glider_4.png}}
% 	\centerline{\includegraphics[scale=0.25]{morley_glider_5.png}}
% 	\centerline{\includegraphics[scale=0.25]{morley_glider_6.png}}
% 	\centerline{\includegraphics[scale=0.25]{morley_glider_7.png}}
%\caption{The final 3 updates of a glider pattern in the B368/S245 set of rules. Moore neighborhood values are inset in each cell, with birth and survival denoted in purple and ``death" transitions denoted in pinkish yellow.} 
%\label{fig:morley1}
%\end{figure}

\section{Related Work}

Carle's Game combines the fields of cellular automata, open-ended machine learning, and evolutionary computation. The growth and dynamics of cellular automata have been extensively studied \cite{wolfram1983, schiff2005, eppstein2010, wolfram1984}, and thanks to the characteristics of complexity and universal computation of many CA \cite{wolfram1984, cook2004, conwaylifeforum}, CA make a good substrate for modeling physical phenomena \cite{toffoli1987}. Recent work in differentiable neural CA offer interesting possibilities for modern machine learning \cite{mordvintsev2020, randazzo2020, niklasson2021, randazzo2021}. 

Open-endedness is often a desirable trait in artificial life and evolution simulations. Evolutionary computation applied to generic reinforcement learning (RL) tasks exhibits the same phenomena of specification gaming and reward hacking as seen in conventional RL algorithms. For example, with no costs associated with individual height, evolutionary selection for body plans and policies on the basis of forward travel often discovers the unintended solution of being very tall and falling over \cite{ha2019, sims1994a, sims1994c, lehman2020}. A more open-ended approach can limit the reward function surface that can be exploited, at least in principle, but the challenge of producing interesting results remains substantial. A system can apparently meet all the generally accepted traits of open-ended evolution and still produce uninteresting results \cite{hintze2019}.

Previous efforts to build artificial worlds include individual machine code programs competing for computational resources \cite{ray1991, ofria2005}, simulated creatures in physics simulators \cite{sims1994a} \cite{sims1994c} \cite{yaeger1997} \cite{spector2007}, games \cite{grbic2020, channon1997, suarez2019, soros2017}, or abstract worlds \cite{soros2014}. I find the open-ended universes based on cellular automata such as SmoothLife, Lenia, and genelife \cite{rafler2011, chan2019, mccaskill2019} particulary interesting. CA simulations are an attractive substrate for open-ended environments because they are not based on a facsimile of the physics of our own universe, nor are they an ad-hoc or abstract landscape. Instead they have their own internally consistent rules, analagous to physical laws in myriad alternative universes. 

Unlike experiments in Lenia, SmoothLife, or genelife, in Carle's Game agents interact with a CA universe, but they are not of it. Instead the role of agents is more similar to that of the decades-long exploration of Life-like cellular automata by human experimenters. CARLE also has the ability to simulate any of the 262,144 possible Life-like rules, of which many exhibit interest patterns of growth and decay and/or Turing completeness \cite{eppstein2010, conwaylifeforum}. %This flexibility allows for general and adaptive agents to be developed, as meta-learning agents can be trained under several different sets of rules and evaluated in others. 

Another possible advantage of CARLE comes from the simplicity of CA universes combined with execution on a graphics processing unit (GPU). CARLE is written with PyTorch \cite{paszke2019}, and the implementation of CA rules in the matrix multiplication and convolutional operations, operations that have been well-optimized in modern deep learning frameworks, makes hardware acceleration straightforward. Although CARLE does not take advantage of CA-specific speedup strategies like the HashLife algorithm \cite{gosper1984}, it uses vectorization and GPU acceleration to achieve in excess of 20,000 updates of a 64 by 64 cell grid per second. Combined with the expressiveness and versatility of Life-like cellular automata, it is my hope that these characteristics will make CARLE an enabling addition to available programs for investigating machine creativity and exploration.

\section{Carle's Game and the Cellular Automata Reinforcement Learning Environment}

% describe carle

\subsection{What's Included}

% first mention of CARLE in the text, needs acronym explained
Carle's Game is built on top of CARLE, a flexible simulator for Life-like cellular automata written in Python \cite{vanrossum2009} with Numpy \cite{harris2020} and PyTorch. As CARLE is formulated as a reinforcement learning environment, it returns an observation consisting of the on/off state of the entire CA grid at each time step, and accepts actions that specify which cells to toggle before applying the next Life-like rule updates. The action space is a subset of the observation space and the (square) dimensions of both can be user-specified, with default values of an observation space of 128 by 128 cells and an action space of 64 by 64 cells. Although both are naturally 2D, they are represented as 4-dimensional matrices (PyTorch Tensors), {\itshape i.e.} with dimensions $Nx1xHxW$, or number of CA grids by 1 by height by width. If CARLE receives an action specifying every cell in the action space to be toggled, the environment is reset and all live cells are cleared. In addition to the CA simulation environment, CARLE with Carle's Game includes: 

\begin{itemize}
	\item{ {\bfseries Reward Wrappers: Growth, Mobiliy, and Exploration Bonuses.} CARLE always returns a reward of 0.0, but several reward wrappers are provided as part of CARLE and Carle's Game to provide motivation for agents to explore and create interesting patterns. Implemented reward wrappers include autoencoder loss and random network distillation exploration bonuses \cite{burda2018}, a glider/spaceship detector, and a reward for occupying specific regions of the CA universe. }
	\item{ {\bfseries Starter Agents: HARLI, CARLA, and Toggle.} I have included several agent policies as a starting point for developing innovative new policies. These shouldn't be expected to be ideal architectures and experimenters are encouraged to explore. Starter agents include Cellular Automata Reinforcement Learning Agent (CARLA), a policy based on continuous-valued cellular automata rules, and HARLI (Hebbian Automata Reinforcement Learning Improviser), a policy again implemented in continuous-valued neural CA that learns to learn by optimizing a set of Hebbian plasticity rules. Finally, Toggle is an agent policy that optimizes a set of actions directly, which are only applied at the first step of simulation.}
	\item{ {\bfseries Starter Algorithm: Covariance Matrix Adapataion Evolution Strategy (CMA-ES).} Carle's Game includes an implementation of CMA-ES \cite{hansen2016} for optimizing agent policies.}
	\item{ {\bfseries Human-Directed Evolution.} In addition to reward wrappers that can be applied to CARLE, Carle's Game includes interactive evolution (implemented in a Jupyter notebook) to optimize agent policies with respect to human preferences.}
\end{itemize}

\subsection{Performance}

\begin{figure}[tb]
	\centerline{\includegraphics[scale=0.25]{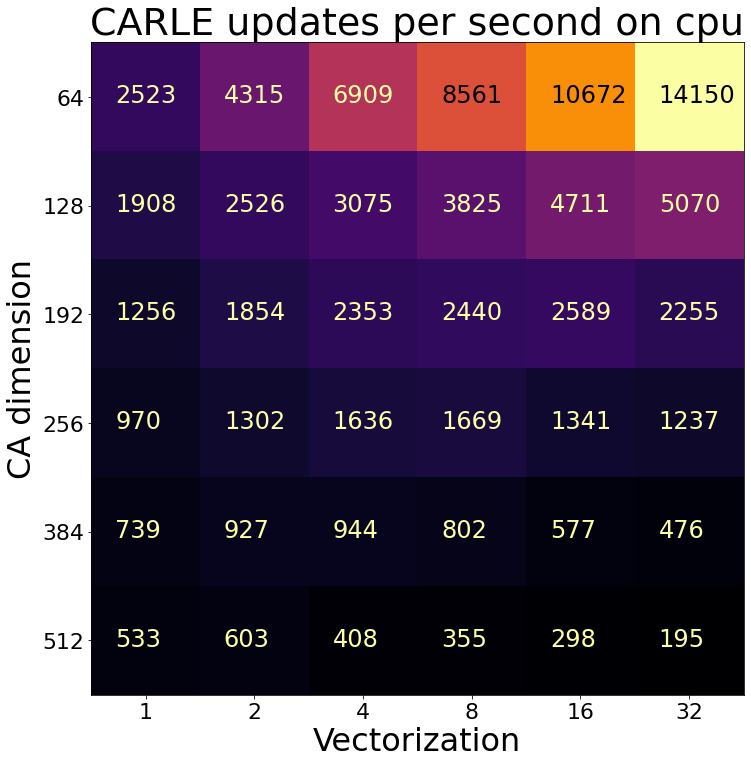}}
\caption{CARLE performance running on a 24-core CPU, measured in CA grid updates per second.}
\label{fig:cpu_performance}
\end{figure}

\begin{figure}[tb]
	\centerline{\includegraphics[scale=0.25]{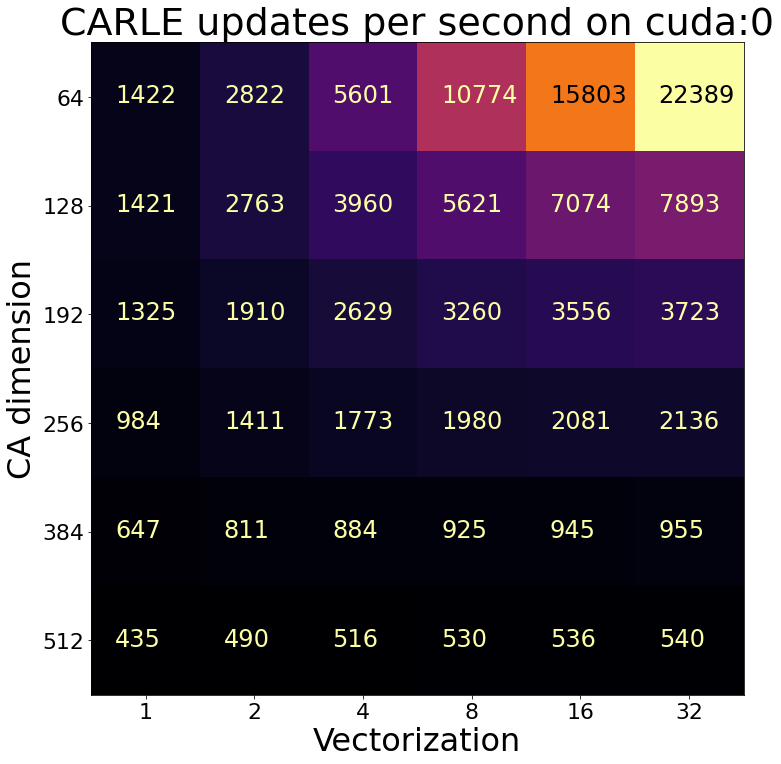}}
\caption{CARLE performance running with device set to an Nvidida GTX 1060 GPU, measured in CA grid updates per second.}
\label{fig:cuda_performance}
\end{figure}

CARLE utilizes a combination of environment vectorization and GPU acceleration to achieve fast run times. On a consumer desktop with a 24-core processor (AMD 3960x Threadripper) and an Nvidia GTX 1060 GPU, CARLE runs at a speed of more than 22,000 updates per second of a 64 by 64 grid running Game of Life (B3/S23), amounting to about 90 million cell updates per second. Additional performance gains are likely available with lower precision operations with modern (RTX) GPU acceleration. To assess CARLE performance on other systems, I've provided the Jupyter notebook I used to create Figures \ref{fig:cpu_performance} and \ref{fig:cuda_performance}, available at \url{https://github.com/rivesunder/carles_game}

\subsection{Reward Wrappers}

The reward hypothesis, also sometimes known as the reinforcement learning hypothesis, has a somewhat ambiguous attribution \cite{suttonreward, littmanreward}, but no matter who described the idea first it can be summarized as follows: intelligent behavior arises from an agent seeking to maximize the rewards it receives from its environment, or at least an agent's goals can be fully described as an effort to maximize cumulative rewards. Whether this is reflective of real-world intelligence is a discussion for another time, but it does mean that open-ended environments returning no nominal reward are a challenge for most RL algorithms. 

While CARLE is formulated as a reinforcement learning environment and uses the \texttt{observation, reward, done, info = env.step(action)} Gym API \cite{brockman2016}, the environment always returns a reward value of 0.0. Ultimately Carle's Game uses crowd-sourced voting to evaluate user submissions, and judging machine creativity and exploration is likely to be just as subjective as judging the creative outputs of human artists and explorers. Designing a system of derived rewards, ideally in such a way as to instill an internal sense of reward for exploration, is a principal aspect of the Carle's Game challenge. I have included several examples of potential reward wrappers that intervene between the universe of CARLE and agents interacting with it to add rewards for agents to learn from. These may be used directly and in combination to generate reward proxies or serve as a template and inspiration for custom reward wrappers. 

\begin{figure}[tb]
	\centerline{\includegraphics[scale=0.2]{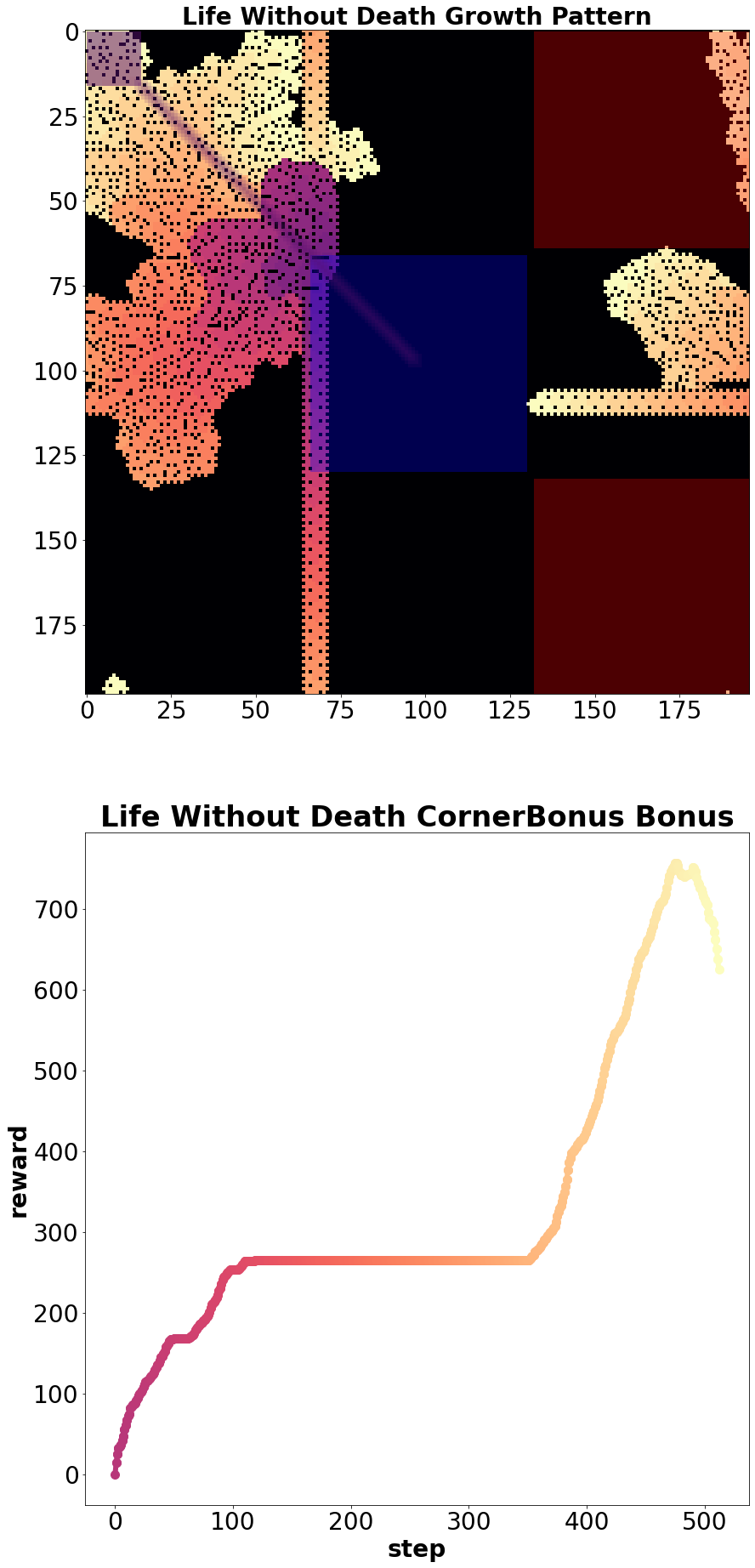}}
	\caption{A manually designed growth pattern that yields a positive cumulative reward from the CornerBonus wrapper (B3/S012345678 rules). The areas tinged in red generate a -1.0 reward for each live cell, the area tinged in purple returns a +1.0 reward for every live cell, and the area tinged in blue represents the action space.}
\label{fig:cornerbonus}
\end{figure}

Carle's Game includes two implementation of exploration bonus rewards, one based on autoencoder loss and the other on random network distillation \cite{burda2018}. It also includes wrappers designed to reward moving machines such as spaceships and gliders and another that grants a reward for occupying specific cells in the top left corner. The corner bonus wrapper (Figure \ref{fig:cornerbonus} yields negative reward for live cells in the right hand corners and a positive reward for live cells in top left corner, and along the diagonal of the grid universe between the action space and the top left corner. The speed reward wrapper calculates the change between the center of mass of all live cells between time steps, and while machine exploration and crowd evaluation of submissions are open-ended and subjective, developing agents that can re-discover known patterns-of-interest like spaceships and gliders provides an attractive first step that parallels early human exploration of Conway's Life.  

%Machine learning agents finding themselves a random initialization of parameter values will

\begin{figure}[tb]
	\centerline{\includegraphics[scale=0.2]{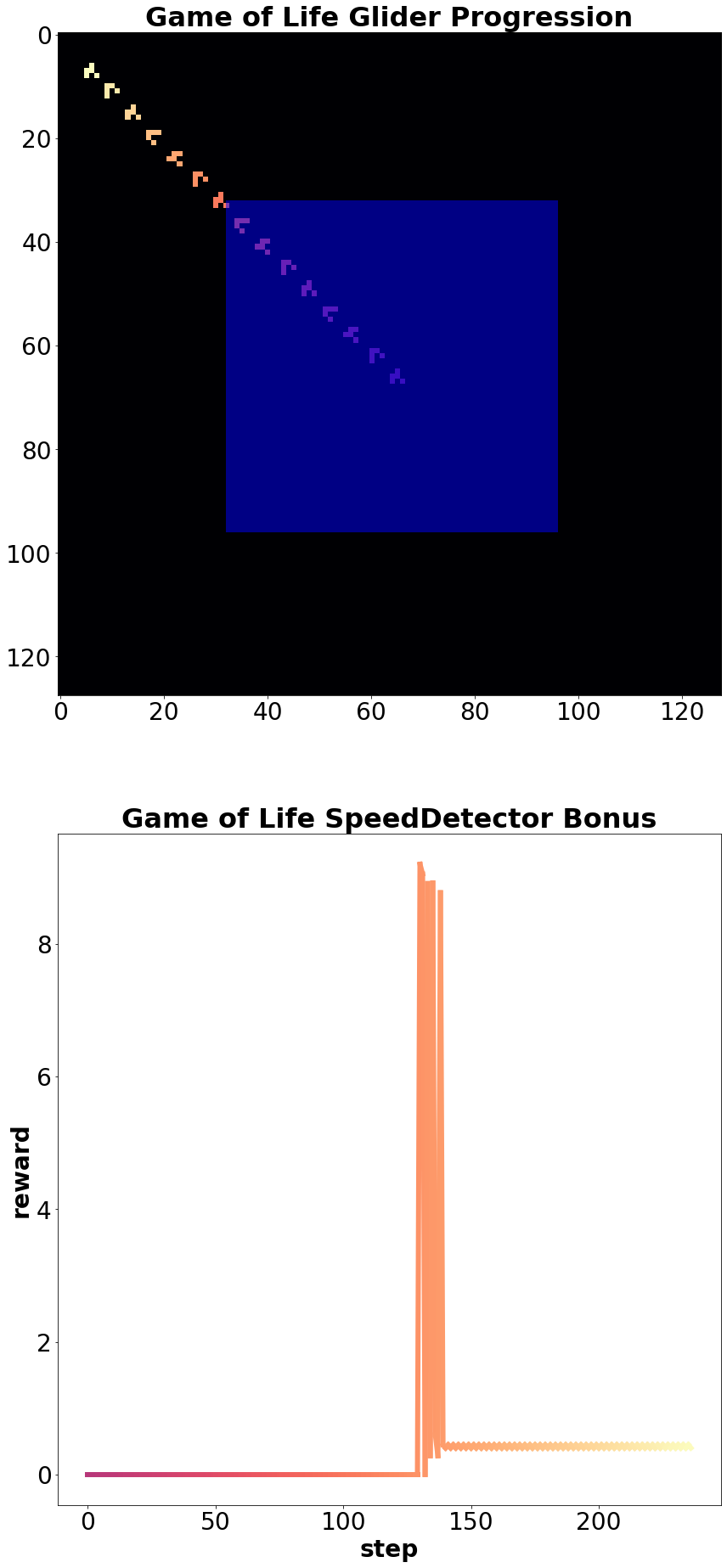}}
	\caption{Center of mass speed reward applied to a Game of Life glider. This reward wrapper ignores the cells in the action area (tinged in blue) when calculating the center of mass, resulting in a large reward spike when live cells first escape the action space, readily apparent in the reward curve. In this example, a Game of Life glider travels across the CA grid. After the glider exits the action space, the reward oscillates around the average speed of the object, reflecting the periodic diagonal motion of the glider.}
\label{fig:speeddetector}
\end{figure}

An example of the speed-based reward wrapper applied to a Life glider is shown in Figure \ref{fig:speeddetector}. This reward system considers cells inside of the action space to all have a central location of (0,0), leading to a large reward spike when an object first leaves the area available to agent manipulation. As CARLE simulates a toroidal CA universe, another reward spike occurs (not shown in Figure \ref{fig:speeddetector}) when an object leaves by one edge and returns by another. This boundary crossing advantage was exploited by agents in experiments described in section \ref{exp}.

%\begin{figure}[tb]
%	\centerline{\includegraphics[scale=0.2]{autoencoder_reward.png}}
%\caption{Autoencoder loss based exploration. The colored background stripes differentiate the puffer pattern's progression at different timesteps separated by 100 grid updates. The update schedule for the autoencoder is visible in the ``stair-step" effect of the reward curve. The autoencoder is a fully convolutional model, and as a result the rewards here have translational equivariance.}
%\label{fig:autoencoder}
%\end{figure}
%
%\begin{figure}[tb]
%	\centerline{\includegraphics[scale=0.2]{randomnetworkdistillation_reward.png}}
%\caption{Exploration reward based on random network distillation, where a prediction model attemps to learn the input output mapping of a random network. The puffer object is shown at each 100th step of its progression with a different background color. Unlike the autoencoder loss reward, the implementation of random network distillation in Carle's Game includes a fully-connected neural network layer at the output, allowing the reward system to be ``surprised" when the puffer disappears of the edge of the of the CA grid and re-appears on the opposite side.}
%\label{fig:rnd}
%\end{figure}

\begin{figure}[tb]
	\centerline{\includegraphics[scale=0.2]{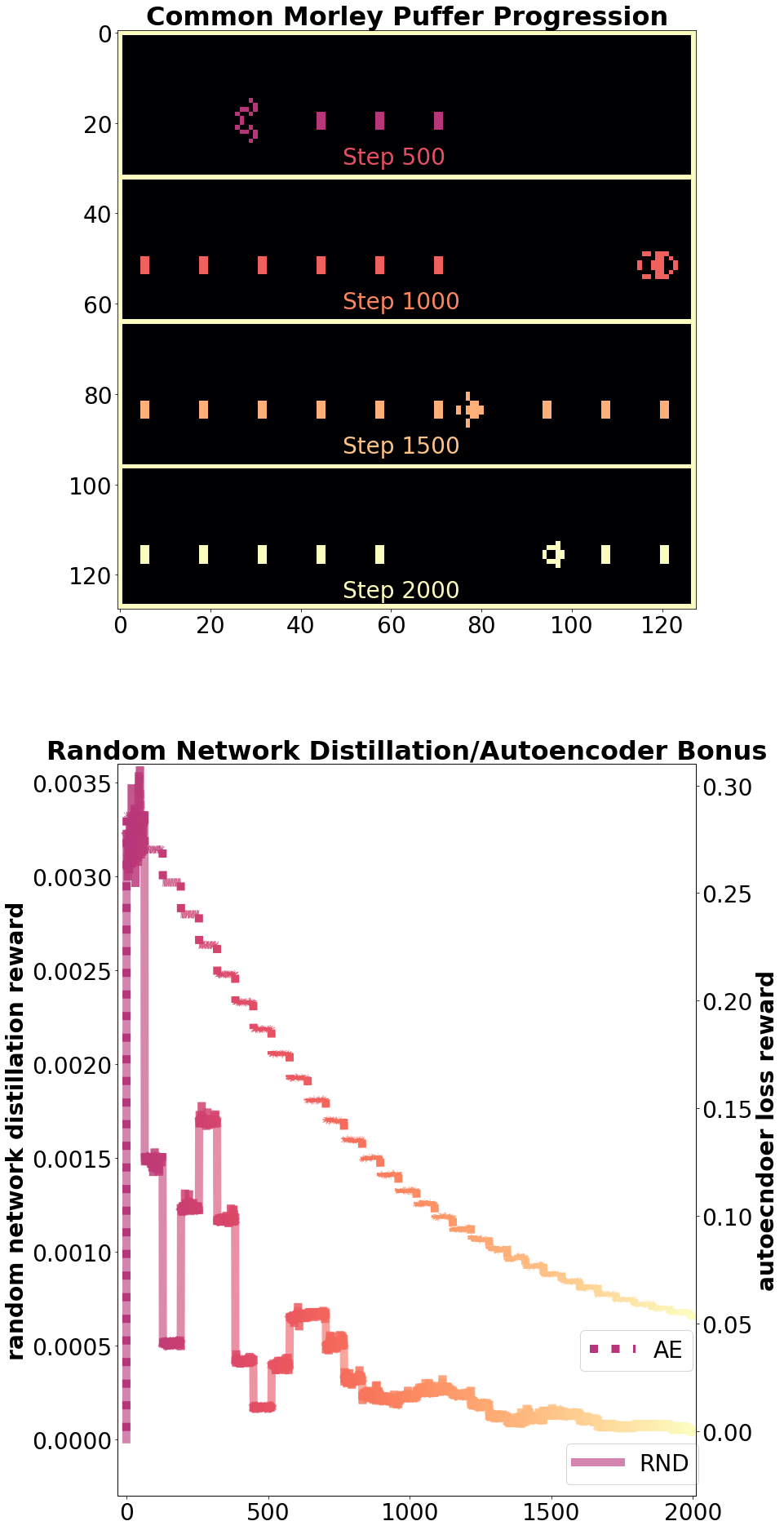}}
\caption{Exploration rewards based on random network distillation and autoencoder loss. In random network distillation, a prediction model attemps to learn the input-output mapping of a random network, while autoencoder reward is based on the autoencoder loss of reconstructing environment observations directly. Snapshots of the progression of a puffer object are displayed with a color map corresponding to the color of the corresponding step in the reward curves, lighter coloration indicates later time steps. Note the different scales for random network distillation and autoencoder based rewards. Unlike the autoencoder loss reward, the implementation of random network distillation in Carle's Game includes a fully-connected neural network layer at the output, allowing the reward system to be ``surprised" and generate higher reward when the puffer disappears of the edge of the of the CA grid and re-appears on the opposite side.}
\label{fig:rnd}
\end{figure}

Autoencoder loss and random network distillation approximate the novelty of a set of observations, as states that are visited less frequently lead to relatively poor performance compared to observations often encountered \cite{burda2018}. These offer slightly different behavior in practice. The autoencoder loss reward wrapper is implemented with a fully convolutional neural network model, and therefore exhibits translation equivariance in its response. Therefore the exploration bonus for a given pattern can be expected to be the same regardless of where on the CA grid the pattern appears. The implementation of random network distillation included with CARLE, on the other hand, includes fully connected neural networks in the final layers of both the random and the prediction network. The fully connected layers do not exhibit the translation equivariance of convolutional layers, so a jump in reward occurs when the Morley puffer leaves via one edge and appears on the opposite side of the grid, visible in Figure \ref{fig:rnd}. Both examples are demonstrated for the same progression of a common puffer in the Morley/move rule set (B368/S245). 

\section{Carle's Game Baselines}

\begin{figure}[tb]
	\centerline{\includegraphics[scale=0.2]{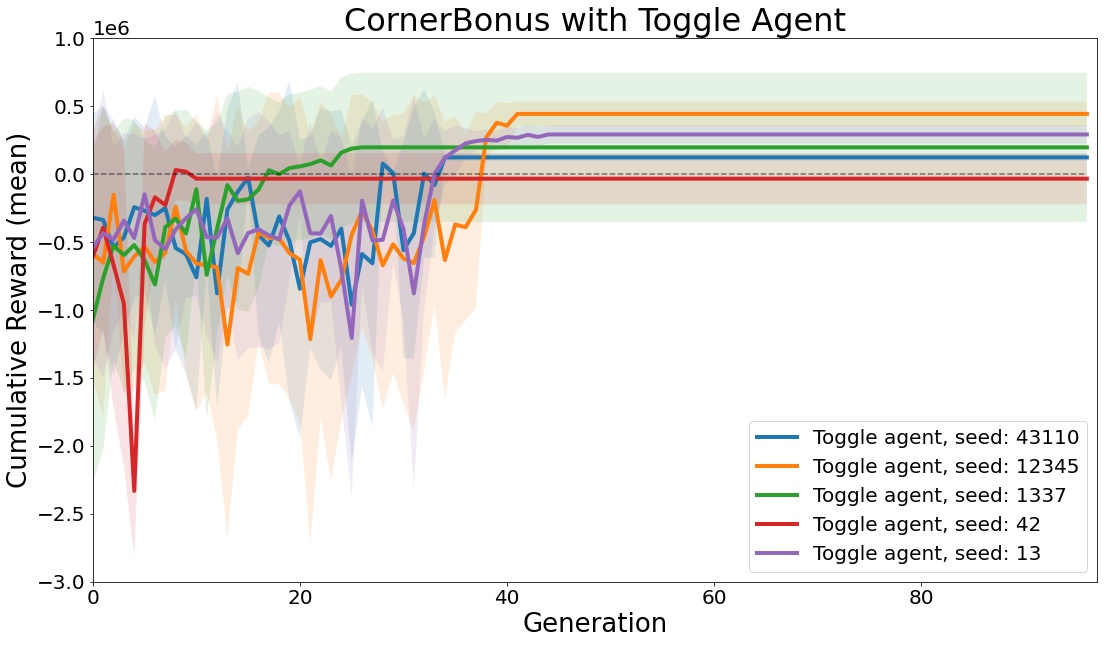}}
\caption{}
\label{fig:corneragent}
\end{figure}

\subsection{Carle's Game Baselines: Discovering the Morley Glider, Corner-Seeking, and Hacking the Mobility Reward}

This section briefly presents CMA-ES training of agents using the corner bonus or a combination of the speed and random network distillation reward wrappers. 

\subsubsection{Reaching for the Corner}

The first experiment used CMA-ES \cite{hansen2016} to optimize patterns directly using the ``Toggle" agent. The training curves are shown for 5 random seeds in Figure \ref{fig:corneragent}. Of the 5 random seeds, 4 experimental runs found a pattern that achieves a positive cumulative reward, {\itshape i.e.} a pattern that for the most part avoids the negative reward corners and occupies at least some of the positive reward cells. This experiment used the B3/S245678 growth set of rules, and an example of a manually designed pattern achieving positive cumulative reward on this task is shown in Figure \ref{fig:cornerbonus}.

\subsubsection{Rediscovering Gliders}

\label{exp}

A second experiment used the combination of random network distillation and a reward based on changing center of mass. This experiment was intended to investigate the ability of agents to learn to create gliders and small spaceships. Patterns were evolved directly using the Toggle agent, as well as dynamic agents CARLA and HARLI. This experiment was conducted using B368/S245 rules.  

\begin{figure}[tb]
	\centerline{\includegraphics[scale=0.3]{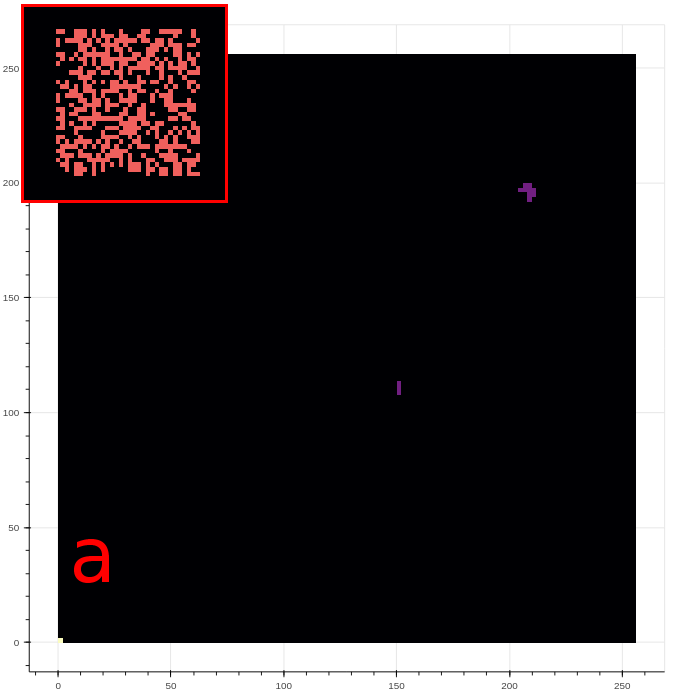}}
	\centerline{\includegraphics[scale=0.3]{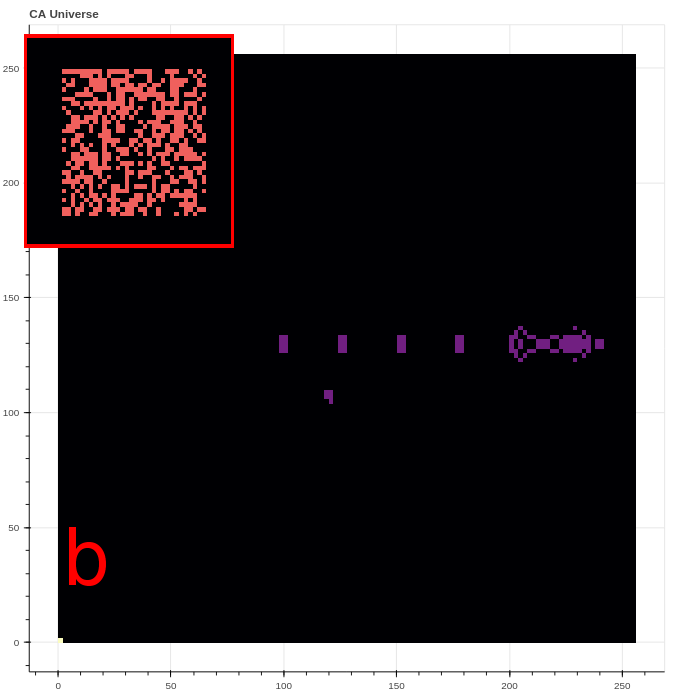}}
	\caption{Patterns discovered by CMA-ES optimization of the Toggle agent. a) Pattern that produces a glider. b) Pattern that produces a puffer. The evolved patterns are shown inset in red.}
\label{fig:toggle_glider}
\end{figure}

Experiments involving the Toggle agent managed to discover several patterns that produce gliders. It's worth noting that in preliminary experiments optimizing the Toggle agent with the speed bonus directly, evolution runs with small population sizes and no exploration bonuses often got stuck in local optima and failed to find mobile patterns. Figure \ref{fig:toggle_glider} demonstrates the discovery of a pattern that produces a glider, and another that produces a puffer. 

\begin{figure}[tb]
	\centerline{\includegraphics[scale=0.2]{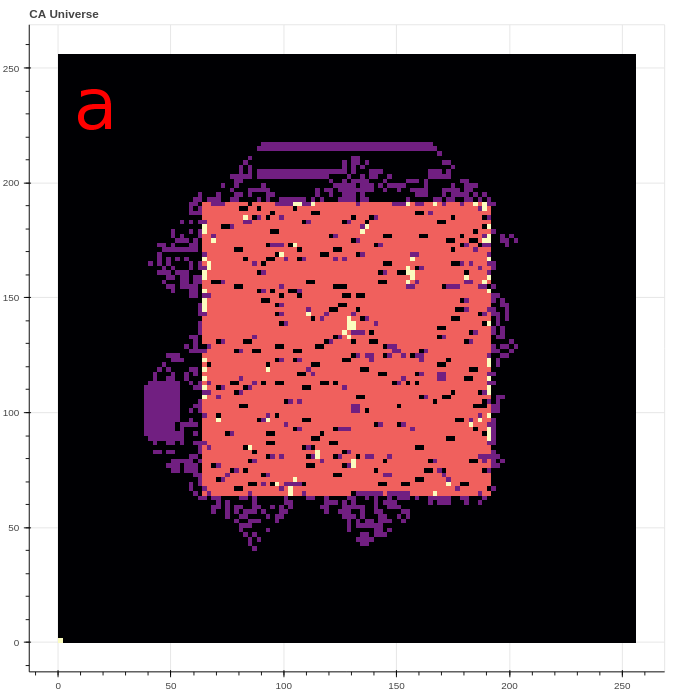}}
	\centerline{\includegraphics[scale=0.2]{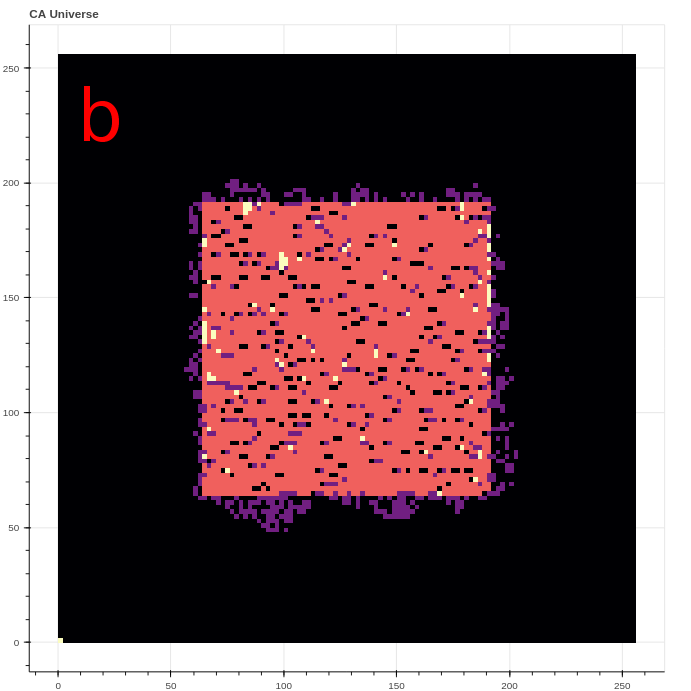}}
	\caption{Speed + RND reward strategies employed by CA-based agents CARLA and HARLI (CARLA examples shown here). a) Wave strategy: activating a line of cells at the action space boundaries creates a wave moving away from the action area, exploiting the center of mass calculations that consider all cells within the action space to be at $(0,0)$ b Chaotic boundary strategy: by continuously activating cells near the boundary, live cells continuously appear and disappear just outside of the action. So long as the live cells don't perfectly balance each other out, this results in a changing center of mass and a continuous stream of small rewards. Unlike the Toggle agent, CARLA and HARLI are dynamic and can continuously perturb the action space, and are not constrained to learn to produce a persistent mobile pattern. }
\label{fig:carla_glider}
\end{figure}

Perhaps unsurprisingly, the active agents trained in the speed reward experiment found surprising and unintended strategies to generate high rewards. The three strategies include resetting the environment, generating a ``wave", or maintaining a ``chaotic boundary". The reset strategy generates a high reward by clearing all live cells from the CA grid, effectively moving their center of mass from some finite value to the center of the grid. The wave strategy is often seen in the first few steps, and takes advantage of the ``B3" rule contribution to produce a moving line of live cells visible in Figure \ref{fig:carla_glider}a. The chaotic boundary strategy takes advantage of the jump in reward produced when live cells transiently appear just outside of the action space.

\begin{figure}[tb]
	\centerline{\includegraphics[scale=0.3]{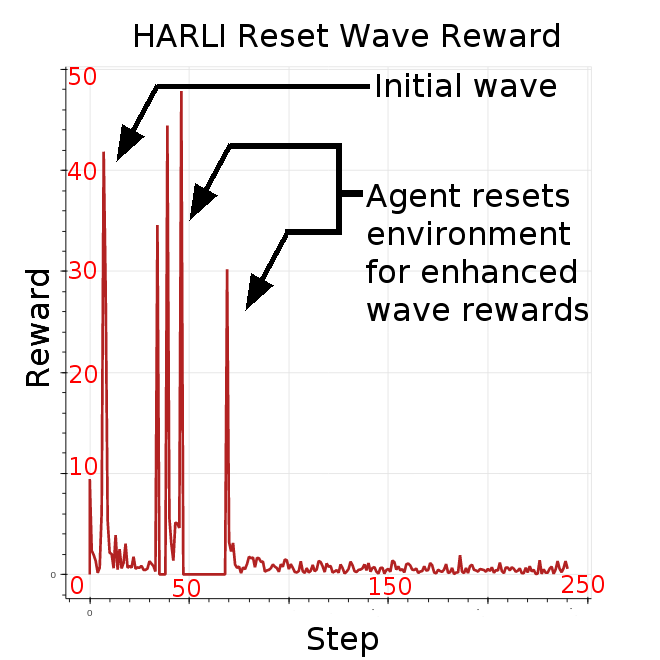}}
	\caption{The HARLI agent continuously modifies its parameters according to Hebbian rules with evolved weights. Crucial time points related to the environment reset strategy are called out with black arrows. Subsequent steps use the chaotic boundary strategy for a continuous stream of small rewards.}	
\label{fig:harli_exploit}
\end{figure}

While HARLI and CARLA agents both exploit similar wave and chaotic boundary strategies, HARLI also is capable of resetting the environnment by setting all toggles to 1. The reset and wave strategy used together generate extremely high average rewards that may be 10 times greater than a common Life glider. An interactive demonstration of the reset-wave strategy at several evolutionary snapshots (trained with 4 different B3/Sxxx rules) can be found at \url{https://github.com/riveSunder/harli_learning}, and an example of this strategy under B368/S245 rules is presented in the reward plot in Figure \ref{fig:harli_exploit}.

\section{Concluding Remarks}

As introduced in this paper, CARLE along with Carle's Game provides a fast and flexible framework for investigating machine creativity in open-ended environments. Optimizing initial patterns directly using the Toggle policy managed to re-discover both the glider and the common puffer in the B368/S245 Morley/Move rules. We also discussed two agents based on continuous-valued CA policies, which both managed to find effective (and frustratingly unintended) strategies for garnering rewards from a center of mass speed reward wrapper. These unintended strategies, which can be described as reward hacking exploits, underscore the challenge of providing motivation to machine learning agents in the face of complexity.  

The environment CARLE is available at \url{https://github.com/rivesunder/carle} and the Carle's Game challenge, including the code used to produce the experiments and figures described in this article, is maintained at \url{https://github.com/rivesunder/carles_game}. I intend to actively maintain these projects at least until the IEEE Conference on Games in August 2021, where Carle's Game is among the competition tracks \cite{ieeecog}, and the code is permissively licensed under the MIT License.

%\vspace{12pt}
%\color{red}
%IEEE conference templates contain guidance text for composing and formatting conference papers. Please ensure that all template text is removed from your conference paper prior to submission to the conference. Failure to remove the template text from your paper may result in your paper not being published.

\end{document}